\documentclass{article}

\usepackage[final]{corl_2019}      

\newcommand{\todo}[1]{\textcolor{blue}{TODO: #1}}

\newcommand\lblsec[1]{\label{sec:#1}}

\newcommand\lblfig[1]{\label{fig:#1}}
\newcommand\reffig[1]{Figure \ref{fig:#1}}
\newcommand\lbltbl[1]{\label{tbl:#1}}
\newcommand\reftbl[1]{Table \ref{tbl:#1}}

\usepackage{tango}
\usepackage{wrapfig}

\usepackage{amssymb}
\usepackage{amsmath}
\usepackage{multirow}
\usepackage{booktabs}
\usepackage{siunitx}
\usepackage{tabularx}
\usepackage{graphicx}
\usepackage{tikz}
\usepackage{pgfplots}
\usepackage{subcaption}
\usetikzlibrary{positioning}
\usetikzlibrary{arrows}
\usetikzlibrary{calc,fit}
\usepackage{dblfloatfix}
\usepackage[title]{appendix}

\usepackage{verbatim}

\newcommand\mypara[1]{\vspace{0mm}\noindent\textbf{#1}}

\DeclareMathOperator*{\minimize}{minimize}

\usepackage[ruled,vlined]{algorithm2e}

\pgfplotsset{
	barplot/.style={
		ybar,
		bar width=12pt,
		width=\textwidth,
    	grid style=solid,
		axis line style={draw=none},
		axis x line=bottom,
		xtick style={draw=none},
    	xticklabel style={font=\footnotesize},
		axis y line=left,
		ytick style={draw=none},
    	yticklabel style={font=\footnotesize},
		ymajorgrids=true,
		legend columns=1,
        legend cell align={left},
    	legend style={font=\footnotesize},
	}
}

\title{Learning by cheating}

\author{
  Dian Chen \\
  UT Austin \\
  \And
  Brady Zhou \\
  Intel Labs, UT Austin \\
  \And
  Vladlen Koltun \\
  Intel Labs \\
  \And
  Philipp Kr{\"a}henb{\"u}hl \\
  UT Austin \\
}

\begin{document}
\maketitle


\begin{abstract}
Vision-based urban driving is hard. The autonomous system needs to learn to perceive the world and act in it. We show that this challenging learning problem can be simplified by decomposing it into two stages. We first train an agent that has access to privileged information. This privileged agent cheats by observing the ground-truth layout of the environment and the positions of all traffic participants. In the second stage, the privileged agent acts as a teacher that trains a purely vision-based sensorimotor agent. The resulting sensorimotor agent does not have access to any privileged information and does not cheat. This two-stage training procedure is counter-intuitive at first, but has a number of important advantages that we analyze and empirically demonstrate. We use the presented approach to train a vision-based autonomous driving system that substantially outperforms the state of the art on the CARLA benchmark and the recent NoCrash benchmark. Our approach achieves, for the first time, 100\% success rate on all tasks in the original CARLA benchmark, sets a new record on the NoCrash benchmark, and reduces the frequency of infractions by an order of magnitude compared to the prior state of the art.
\end{abstract}



\section{Introduction}

How should we teach autonomous systems to drive based on visual input?
One family of approaches that has demonstrated promising results is imitation learning~\cite{Argall2009,Osa2018}.
The agent is given trajectories generated by an expert driver, along with the expert's sensory input. The goal of learning is to produce a policy that will mimic the expert's actions given corresponding input~\cite{pomerleau1989alvinn,lecun2005off,bojarski2016end,codevilla2018end,Pan2018AgileAD}.

Despite impressive progress, learning vision-based urban driving by imitation remains hard. The agent is tasked with organizing the ``blooming, buzzing confusion" of the visual world~\cite{James1890} by correlating it with a set of actions shown in the demonstration. A recent study argues that even with tens of millions of examples, direct imitation learning does not yield satisfactory driving policies~\cite{bansal2018chauffeurnet}.

In this paper, we show that imitation learning for vision-based urban driving can be made much more effective by decomposing the learning process into two stages. First, we train an agent that has access to privileged information: it can directly observe the layout of the environment and the positions of other traffic participants. This privileged agent is trained to imitate the expert trajectories. In the second stage, a sensorimotor agent that has no privileged information is trained to imitate the privileged agent. The privileged agent cheats by accessing the ground-truth state of the environment for both training and deployment. The final sensorimotor agent doesn't: it only uses visual input from legitimate sensors (a single forward-facing camera in our experiments) and does not use any privileged information.

The effectiveness of this decomposition is counter-intuitive. If direct imitation learning -- from expert trajectories to vision-based driving -- is hard, why is the decomposition of the learning process into two stages, both of which perform imitation, any better?

\begin{figure}[t]
    \centering
    \begin{subfigure}[b]{0.5\textwidth}
      \raggedright
      \includegraphics[width=0.9\textwidth,page=1]{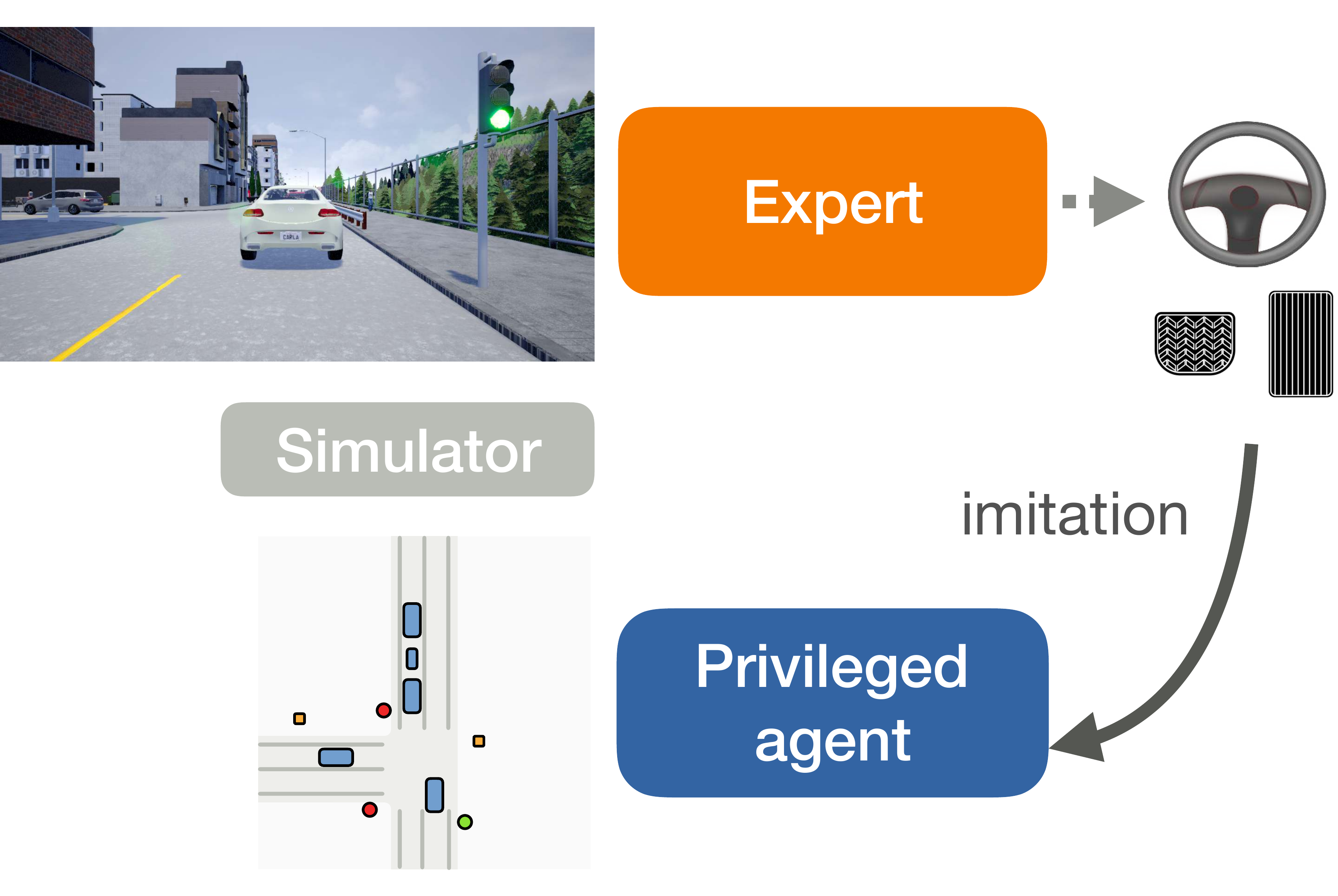}
      \caption{Privileged agent imitates the expert}
    \end{subfigure}%
    \begin{subfigure}[b]{0.5\textwidth}
      \raggedleft
      \includegraphics[width=0.9\textwidth,page=2]{figures/teaser.pdf}
      \caption{Sensorimotor agent imitates the privileged agent}
    \end{subfigure}
    \caption{Overview of our approach. \textbf{(a)}~An agent with access to privileged information learns to imitate expert demonstrations. This agent learns a robust policy by cheating. It does not need to learn to see because it gets direct access to the environment's state. \textbf{(b)}~A sensorimotor agent without access to privileged information then learns to imitate the privileged agent. The privileged agent is a ``white box" and can provide high-capacity on-policy supervision. The resulting sensorimotor agent does not cheat.}
    \label{fig:teaser}
    \vspace{-3mm}
\end{figure}

Conceptually, direct sensorimotor learning conflates two difficult tasks: learning to see and learning to act. Our procedure tackles these in turn. For the privileged agent, in the first stage, perception is solved by providing direct access to the environment's state, and the agent can thus focus on learning to act. In the second stage, the privileged agent acts as a teacher, and provides abundant supervision to the sensorimotor student, whose primary responsibility is learning to see. This is illustrated in Figure~\ref{fig:teaser}.

Concretely, the decomposition provides three advantages. First, the privileged agent operates on a compact intermediate representation of the environment, and can thus learn faster and generalize better~\cite{wang2019monocular,Zhou2019scirob}. In particular, the representation we use (a bird's-eye view) enables simple and effective data augmentation that facilitates generalization.

Second, the trained privileged agent can provide much stronger supervision than the original expert trajectories. It can be queried from any state of the environment, not only states that were visited in the original trajectories. This enables automatic DAgger-like training in which supervision from the privileged agent is gathered adaptively via online rollouts of the sensorimotor agent~\cite{ross2011reduction,Sun2017,Pan2018AgileAD}. It turns passive expert trajectories into an online agent that can provide adaptive on-policy supervision.

The third advantage is that the privileged agent produced in the first stage is a ``white box", in the sense that its internal state can be examined at will. In particular, if the privileged agent is trained via conditional imitation learning~\cite{codevilla2018end}, it can provide an action for each possible command (e.g., ``turn left", ``turn right") in the second stage, all at once, in any state of the environment. Thus all conditional branches of the privileged agent can train all branches of the sensorimotor agent in parallel. In every state visited during training, the sensorimotor student can in effect ask the privileged teacher ``What would you do if you had to turn left here?", ``What would you do if you had to turn right here?", etc. This is both a powerful form of data augmentation and a high-capacity learning signal.

While our training is conducted in simulation -- and indeed relies on simulation in order to access privileged information during the first stage -- the final sensorimotor policy does not rely on any privileged information and is not restricted to simulation. It can be transferred to the physical world using any approach from sim-to-real transfer~\cite{muller2018driving,bewley2018learning}.

We validate the presented approach via extensive experiments on the CARLA benchmark~\cite{dosovitskiy2017carla} and the recent NoCrash benchmark~\cite{codevilla2019exploring}. Our approach achieves, for the first time, 100\% success rate on all tasks in the original CARLA benchmark. We also set a new record on the NoCrash benchmark, advancing the state of the art by 18 percentage points (absolute) in the hardest, dense-traffic condition. Compared to the recent state-of-the-art CILRS architecture~\cite{codevilla2019exploring}, our approach reduces the frequency of infractions by at least an order of magnitude in most conditions.

\mypara{Background.}
Imitation is one of the earliest approaches to learning to drive. It was pioneered by Pomerleau~\cite{pomerleau1989alvinn} and developed further in subsequent work~\cite{lecun2005off,Silver2010navigation,bojarski2016end}. While these earlier investigations focus on lane following and obstacle avoidance, more recent work pushes into urban driving, with nontrivial road layouts and traffic~\cite{codevilla2018end,Sauer2018CORL,liang2018cirl,bansal2018chauffeurnet,codevilla2019exploring}. Our work fits into this line and advances it. In particular, we substantially improve upon the state-of-the-art in recent urban driving benchmarks~\cite{Sauer2018CORL,liang2018cirl,codevilla2019exploring}.

Our work builds on online (or ``on-policy") imitation learning~\cite{ross2011reduction,RossBagnell2014}. Sun et al.~\cite{Sun2017} summarize this line of work and argue that the availability of an optimal oracle can substantially accelerate training. These ideas were applied to off-road racing by Pan et al.~\cite{Pan2018AgileAD}, who trained a neural network policy to imitate a classic MPC expert that had access to expensive sensors. Our work develops related ideas in the context of urban driving.

\section{Method}
\label{sec:method}

The goal of our \emph{sensorimotor agent} is to control an autonomous vehicle by generating steering $s$, throttle $t$, and braking signals $b$ in each time step. The agent's input is a monocular RGB image $I$ from a forward-facing camera, the speed $v$ of the vehicle, and a high-level command $c$ (``follow-lane'', ``turn left", ``turn right", ``go straight")~\cite{codevilla2018end,Sauer2018CORL,liang2018cirl,codevilla2019exploring}. Command input guides the agent along reproducible routes and reduces ambiguity at intersections~\cite{codevilla2018end}.

\begin{figure}[t]
    \centering
    \begin{subfigure}[b]{0.5\textwidth}
        \raggedright
        \includegraphics[width=0.95\textwidth,page=1]{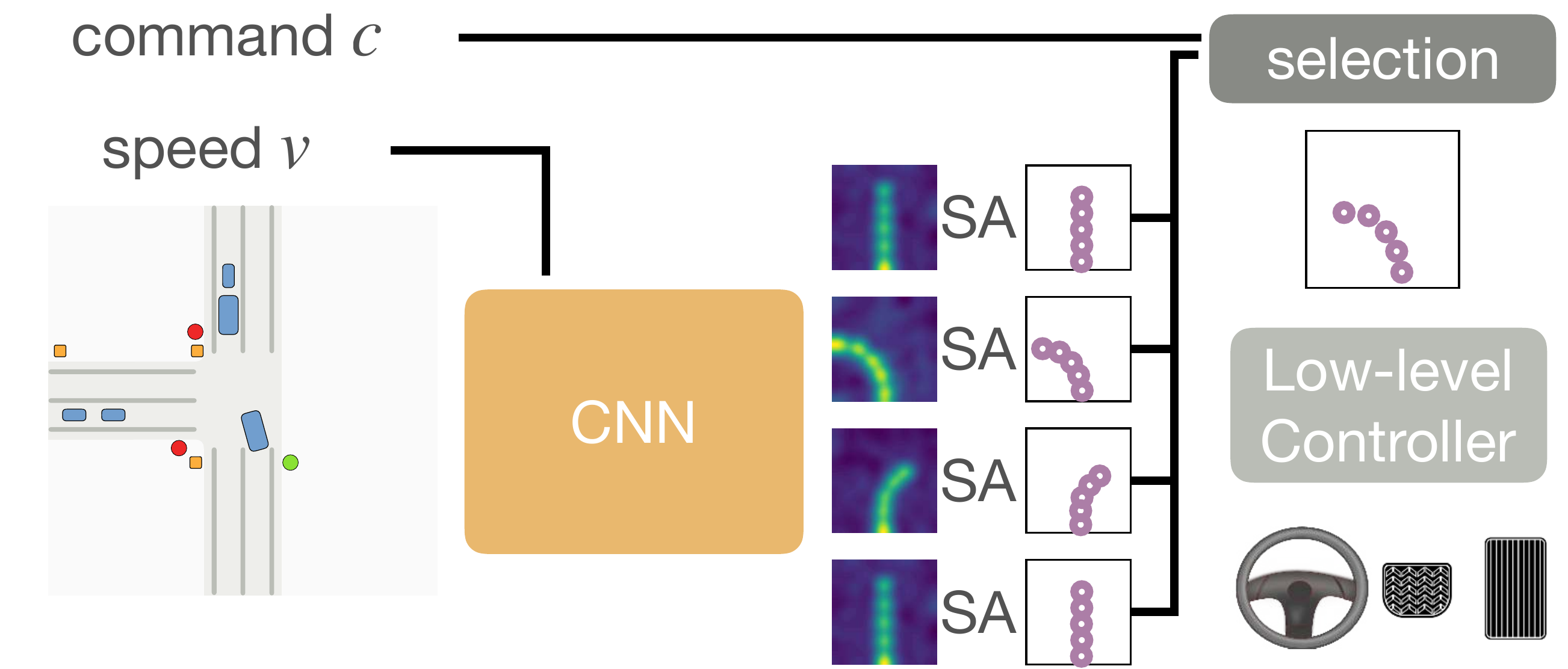}
        \caption{Privileged agent}
        \lblfig{priv_arch}
    \end{subfigure}%
    \begin{subfigure}[b]{0.5\textwidth}
        \raggedleft
        \includegraphics[width=0.95\textwidth,page=3]{figures/pipeline.pdf}
        \caption{Sensorimotor agent}
        \lblfig{sens_arch}
    \end{subfigure}
    \caption{Agent architectures. \textbf{(a)}~The privileged agent receives a bird's-eye view image of the environment and produces a set of heatmaps that go through a soft-argmax layer (SA), yielding waypoints for all commands. The input command selects one conditional branch. The corresponding waypoints are given to a low-level controller that outputs the steering, throttle, and brake. \textbf{(b)}~The sensorimotor agent receives genuine sensory input (image from a forward-facing camera). It produces waypoints in the egocentric camera frame. Waypoints are selected based on the command, projected into the vehicle's coordinate frame, and passed to the low-level controller.}
    \label{fig:agent}
    \vspace{-3mm}
\end{figure}

Our approach first trains a \emph{privileged agent}. This privileged agent gets access to a map $M$ that contains ground-truth lane information, location and status of traffic lights, and vehicles and pedestrians in its vicinity.
This map $M$ is not available to the final sensorimotor agent, and is only accessible by the privileged agent.
Both agents predict a series of waypoints for the vehicle to steer towards.
A low-level PID controller then translates these waypoints into control commands (steering $s$, throttle $t$, brake $b$).
The privileged and sensorimotor agents are illustrated schematically in Figure~\ref{fig:agent}.

Training proceeds in two stages. In the first stage, we train the privileged agent from a set of expert demonstrations.
Next, we train the sensorimotor agent off-policy by imitating the privileged agent on the same set of states as in the first stage, using offline behavior cloning on all command-conditioned branches.
Finally, we train the sensorimotor agent on-policy, using the privileged agent as an oracle that provides adaptive on-demand supervision in any state reached by the sensorimotor student~\cite{ross2011reduction,Sun2017}.

In the following subsections, we explain each aspect of the approach in detail.

\subsection{Privileged agent}

The privileged agent sees the world through a ground-truth map $M \in \{0,1\}^{W \times H \times 7}$, anchored at the agent's current position.
This map contains binary indicators for objects, road features, and traffic lights. See \reffig{mapview}(a) for an illustration.
The task of the privileged agent is to predict $K$ waypoints $\mathbf{w} = \{ w_{1}, \ldots, w_{K} \}$ that the vehicle should travel to.
The agent additionally observes the speed of the vehicle $v$ and the high-level command $c$.
We parameterize this agent ${f^*_\theta : M, v \to \hat{\mathbf{w}}^c}$ as a convolutional network that outputs a series of heatmaps $h^{c,k} \in [0,1]^{W \times H}$, one for each waypoint $k$ and high-level command $c$.
We convert the heatmaps to waypoints using a soft-argmax $\hat{w}^c_k = \sum_{x,y} [x,y]^\top h_{x,y}^{c,k} / \sum_{x,y} h_{x,y}^{c,k}$.
The convolutional network and soft-argmax are end-to-end differentiable.
This representation has the advantage that the input $M$ and the intermediate output $h^{c,k}$ are in perfect alignment, exploiting the spatial structure of the CNN.

The privileged agent is trained using behavior cloning from a set of expert driving trajectories $\{\tau_0, \tau_1, \ldots\}$.
For each trajectory $\tau_i=\{(M_0,c_0,v_0,x_0,R_0), (M_1,c_1,v_1,x_1,R_1), \ldots\}$, we store the ground-truth road map $M_t$, high-level navigation command $c_t$, and the agent's velocity $v_t$, position $x_t$, and orientation $R_t$ in world coordinates.
We generate the ground-truth waypoints from future locations of the agent's vehicle $\mathbf{w}_t = \{R_{t}^{-1} (x_{t+1}-x_t), \ldots, R_{t}^{-1} (x_{t+K}-x_t) \}$, where the inverse rotation matrix $R_{t}^{-1}$ rotates the relative offset into the agent's reference frame.
Given a set of ground-truth trajectories and waypoints, our training objective is to imitate the training trajectories as well as possible, by minimizing the $L_1$ distance between the future waypoints and the agent's predictions:
\begin{equation*}
    \minimize_\theta E_{(M, v, c, \mathbf{w}) \sim \tau}\big[ \|\mathbf{w}  - f^*_\theta(M, v)^{c}\|_1 \big].
\end{equation*}
The training data $M$, $v$, $c$, and $\mathbf{w}$ are sampled from an offline dataset of expert driving trajectories $\tau$.

\begin{wrapfigure}{r}{0.55\textwidth}
\centering
\begin{subfigure}[b]{0.5\linewidth}
\centering
\includegraphics[width=0.9\textwidth, page=9]{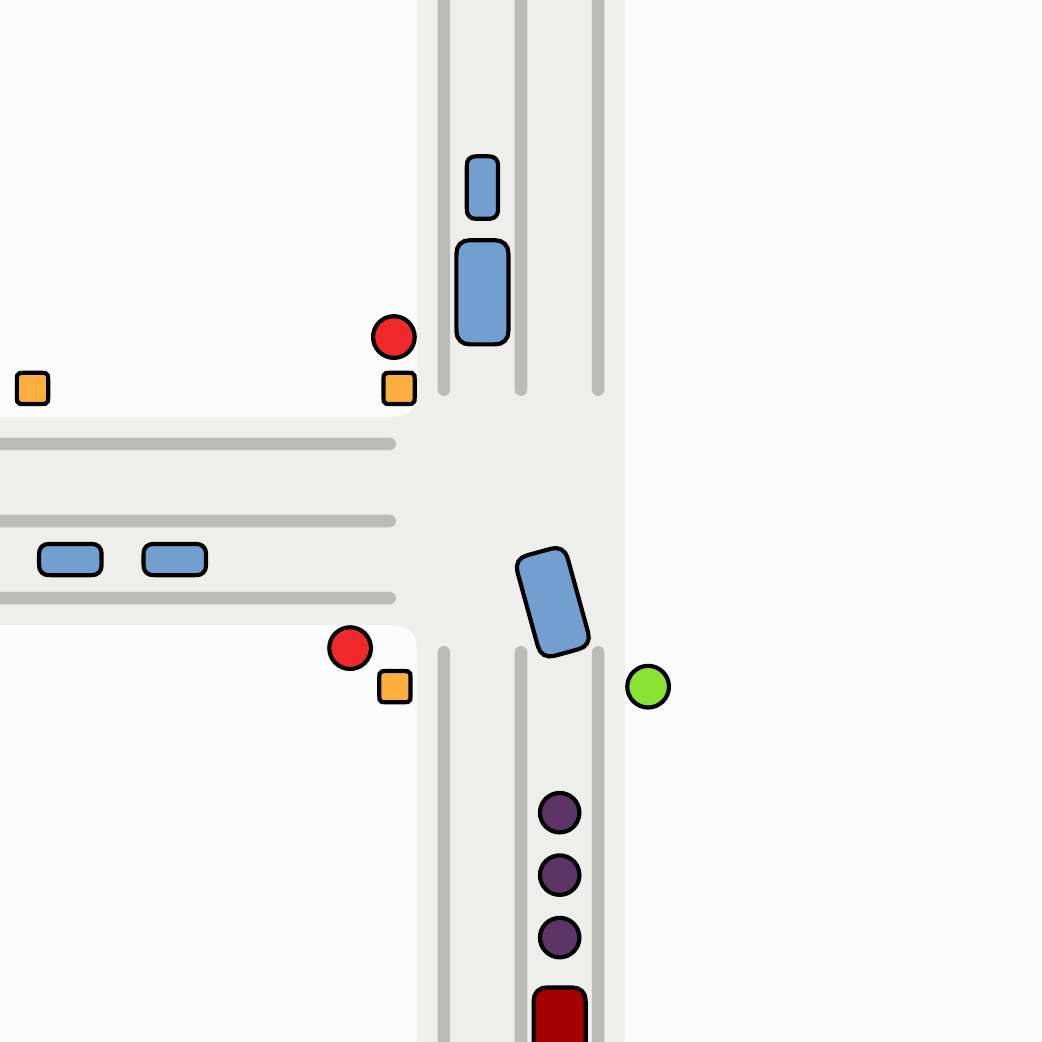}
\caption{Road map}
\end{subfigure}%
\begin{subfigure}[b]{0.5\linewidth}
\centering
\includegraphics[width=0.9\textwidth, page=13]{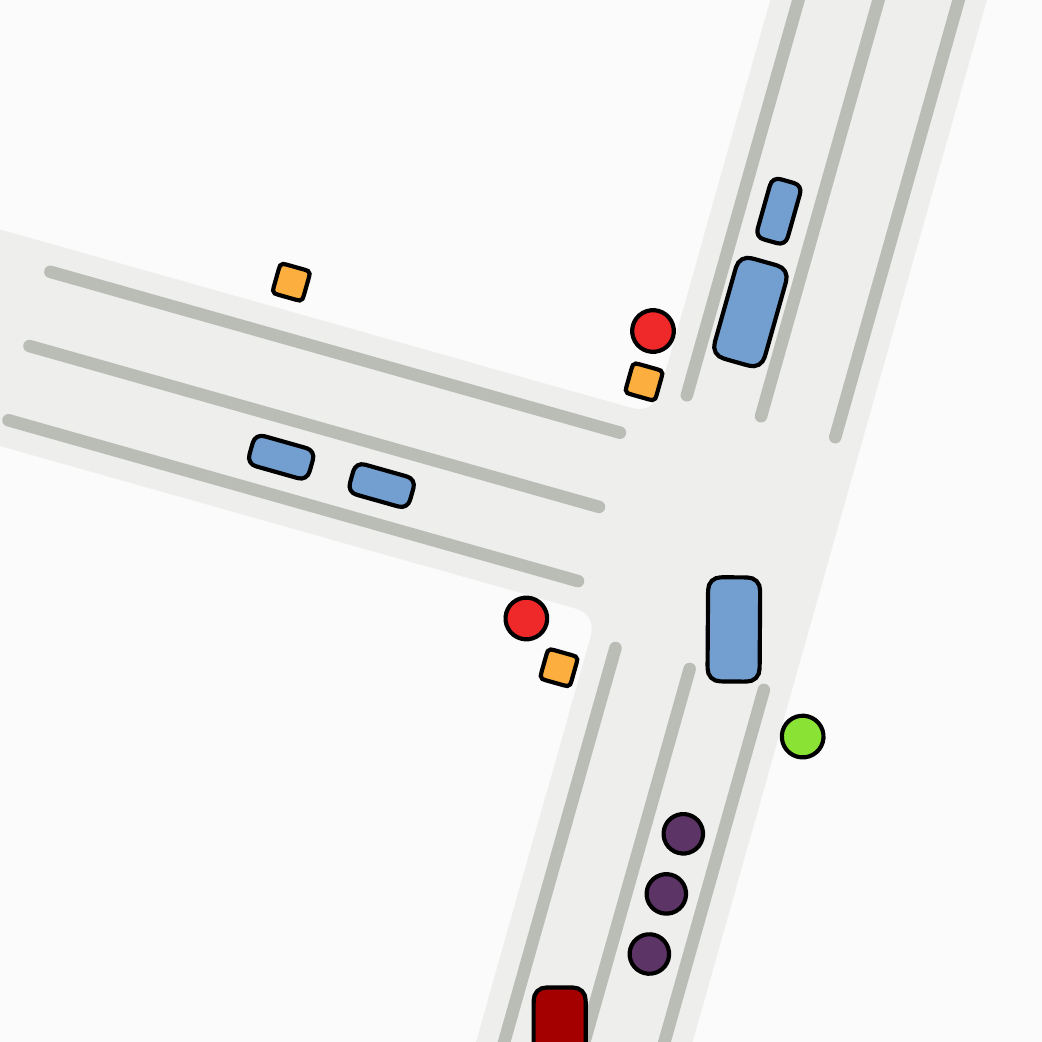}
\caption{Rotation and shift aug.}
\end{subfigure}%
\caption{\textbf{(a)} Map $M$ provided to the privileged agent. One channel each for road (\textcolor{aluminium1}{light grey}), lane boundaries (\textcolor{aluminium2}{grey}), vehicles (\textcolor{skyblue2}{blue}), pedestrians (\textcolor{orange1}{orange}), and traffic lights (\textcolor{chameleon2}{green}, \textcolor{butter2}{yellow}, and \textcolor{scarletred1}{red}). The \textcolor{scarletred3}{agent} is centered at the bottom of the map. The agent's vehicle (\textcolor{scarletred3}{dark red}) and predicted waypoints (\textcolor{plum3}{purple}) are shown for visualization only and are not provided to the network. \textbf{(b)} The map representation affords simple and effective data augmentation via rotation and shifting.}
\lblfig{mapview}

\end{wrapfigure}
In prior imitation learning approaches, data augmentation, namely multiple camera angles~\cite{bojarski2016end} or trajectory noise injection~\cite{laskey2017dart,codevilla2018end,bansal2018chauffeurnet}, was a crucial ingredient.
In our setup, the driving trajectories $\tau$ are noise-free.
We simulate trajectory noise by shifting and rotating the ground-truth map $M$, and propagating the same geometric transformations to the waypoints $\hat{\mathbf{w}}$. This is illustrated in \reffig{mapview}(b). The agent is thus placed in a variety of perturbed configurations (e.g., facing the sidewalk or the opposite lane) and learns to find its way back onto the road by predicting waypoints that lie in the correct lane.
Random rotation and shifting of the map mimic both the multi-camera augmentations and the trajectory perturbations used in other imitation learning setups.
The augmentation is completely offline and does not require any modifications to the data collection procedure or the expert trajectory.

\subsection{Sensorimotor agent}

The structure of the sensorimotor agent $f_\theta : I, v \to \tilde{\mathbf{w}}^c$ closely mimics the privileged agent.
We use a similar network architecture, and the same heatmap and waypoint prediction.
However, the sensorimotor agents only sees the world through an RGB image $I$, and predicts waypoints $\tilde{\mathbf{w}}^c$ in the reference frame of the RGB camera.
This ensures that the input and output representations are aligned.
Waypoints in the map view $\hat{\mathbf{w}}$ and camera view $\tilde{\mathbf{w}}$ are related by a fixed perspective transformation $T_P$ that depends only on the intrinsic parameters and position of the RGB camera, and the size and resolution of the overhead map: $\hat{\mathbf{w}} = T_P(\tilde{\mathbf{w}})$.
We compute this transformation in closed form, as described in the supplement.

The sensorimotor agent is trained to imitate the privileged agent using an $L_1$ loss:
\begin{equation*}
    \minimize_\theta E_{(M, I, v)\!~\sim D}\big[ \|T_p(f(I, v))  - f^*_\theta(M, v)\|_1 \big],
\end{equation*}
where $D$ is a dataset of corresponding road maps $M$, images $I$, and velocities $v$.

This stage has two major advantages.
First, sampling is no longer restricted to the offline trajectories provided by the original expert. In particular, the learning algorithm can sample states adaptively by rolling out the sensorimotor agent during training~\cite{ross2011reduction}.
The second advantage is that the sensorimotor agent can be supervised on all its waypoints and across all commands $c$ at once.
Both of these capabilities provide a significant boost to the driving performance of the resulting sensorimotor agent.

\subsection{Low-level controller}
\lblsec{lowlevel}

The privileged and sensorimotor agents both rely on a low-level controller to translate waypoints into driving commands.
Given a set of waypoints $\hat{\mathbf{w}} = \{\hat{w}_1,\dots,\hat{w}_K\}$ predicted or projected into the vehicle's coordinate frame, the goal of this controller is to produce steering, throttle, and braking commands ($s$, $t$, and $b$, respectively).
We use two independent PID controllers for this purpose.

A longitudinal PID controller tries to match a target velocity $v^*$ as closely as possible.
This target velocity is the average velocity that the vehicle needs to pass through all waypoints:
\begin{equation*}
v_t^* = \frac{1}{K}\sum_{k=1}^{K}\frac{\|\hat{w}_{i} - \hat{w}_{i-1}\|_2}{\delta t},
\end{equation*}
\begin{wrapfigure}{r}{0.33\textwidth}
\vspace{-2em}
\centering
\includegraphics[width=0.25\textwidth,page=1]{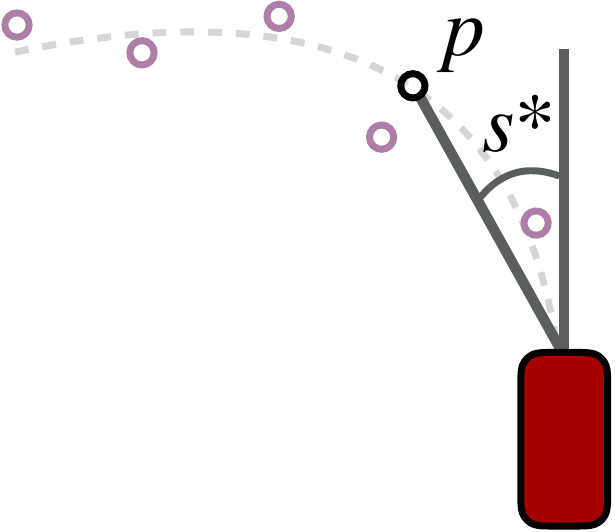}
\caption{Lateral PID controller. Here the agent aims at the projection of the second waypoint onto the fitted arc. $s^*$ denotes the angle between the vehicle and the target point $p$.}
\lblfig{controller}
\vspace{-1em}
\end{wrapfigure}
where $\delta t$ is the temporal spacing between waypoints and ${\hat{w}_0 = [0,0]}$.
The longitudinal PID controller then computes the throttle $t$ to minimize the error $v^* - v$, where $v$ is the current speed of the vehicle.
We ignore negative throttle commands, and only brake if the predicted velocity is below some threshold $v^* < \varepsilon$.
We use $\varepsilon = 2.0\, \textrm{km/h}$.

A lateral PID controller tries to match a target steering angle $s^*$.
Instead of directly steering toward one of the predicted points, we first fit an arc to all waypoints and steer towards a point on the arc, as shown in \reffig{controller}.
This averages out prediction error in individual waypoints.
Specifically, we fit a parametrized circular arc to all waypoints using least-squares fitting.
We then steer towards a point $p$ on the arc. The target steering angle is ${s^* = \tan^{-1}(p_y / p_x)}$.
The point $p$ is a projection of one of the predicted waypoints onto the arc. We use $w_2$ for the straight and follow-the-road commands, $w_3$ for right turn, and $w_4$ for left turn.
Later waypoints allow for a larger turning radius.
These hyperparameters and all parameters of the PID controllers were tuned using a subset of the training routes. 

\section{Implementation details}
\lblsec{implementation}

We implemented the presented approach in PyTorch~\cite{Steiner2019} and both CARLA 0.9.5 and 0.9.6~\cite{dosovitskiy2017carla}.
We render the map $M$ using a slightly modified version of the internal CARLA road map.
We render vehicles, traffic lights, pedestrians, lanes, and road footprint onto separate channels.
We use circles to represent traffic lights.
The map has resolution $320\times 320$ and corresponds to a $64\mathrm{m} \times 64\mathrm{m}$ square in the simulated world.

\mypara{Network architecture.}
\lblsec{arch}
We use a lightweight keypoint detection architecture~\cite{zhou2019objects} to predict waypoints.
The privileged agent uses a randomly initialized ResNet-18 backbone, while the sensorimotor agent uses a ResNet-34 backbone pretrained on ImageNet~\cite{he2016deep}.
Both architectures use three up-convolutional layers to produce an output feature map.
Each up-convolutional layer additionally sees the vehicle velocity $v$ as an input.
The network predicts each waypoint heatmap in a separate output channel using a $1\times 1$ convolution and a linear classifier from a shared feature map.
Following prior work~\cite{codevilla2018end}, the network branches into four heads, where each head produces a $K$-channel heatmap.
The branches represent one of the four high-level commands (``follow-lane", ``turn left", ``turn right", ``go straight").
A differentiable soft-argmax then converts the heatmaps into spatial coordinates.

The input resolution of the privileged agent is $192\times 192$ and the resolution of its output heatmap is $48 \times 48$.
The input image is cropped such that the center of the agent's vehicle is at the bottom of the map.
During training we apply random rotation and shift augmentation to the map.
We first rotate the input image by an angle of $[-5,5]$ degrees uniformly at random, then shift the image left or right by $[-5,5]$ pixels uniformly at random.
This shift corresponds to a $1$m offset in the simulated world.

The sensorimotor agent sees a $384 \times 160$ RGB image as input, and produces $96 \times 40$ heatmaps.
We use the same image augmentations as CIL~\cite{codevilla2018end}, including pixel dropout, blurring, Gaussian noise, and color perturbations. The sensorimotor agent predicts waypoints in camera coordinates, which are then projected into the vehicle's coordinate frame.

\mypara{Training.}
We first train the sensorimotor agent on the same trajectories used to train the privileged agent.
Next, we train the sensorimotor agent online via DAgger~\cite{ross2011reduction}, using the privileged agent as an oracle.
The second stage alone -- without pre-training on the original trajectories -- works equally well in the final accuracy.
However, the online training with DAgger is slower than training on pre-existing trajectories, so we use the first stage to accelerate the overall training process.

We resample the data following \citet{bastani2018verifiable}.
Critical states with higher loss are sampled more frequently.

\section{Results}
\label{sec:results}

We perform all experiments in the open-source CARLA simulator~\cite{dosovitskiy2017carla}.
We train the privileged agent from trajectories of a handcrafted expert autopilot that leverages the internal state of the simulator to navigate through fine-grained hand-designed waypoints.
We collect 100 training trajectories at 10 fps, which amount to about 157K frames (174K in our CARLA 0.9.6 implementation) and 4 hours of driving.
Each frame contains world position $x$, rotation $R$, velocity $v$, monocular RGB camera image $I$, high-level command $c$, and privileged information in the form of a bird's-eye view map $M$.
High-level commands $c$ are computed using a topological graph and simulate a simple navigation system.

Training and validation frames are collected in Town1, under the four training weathers specified by the CARLA benchmark~\cite{dosovitskiy2017carla}.
We use Town2 for the test town evaluation, and do not train or tune any parameters on it.

\mypara{Experimental setup.}
\lblsec{experiment_settings}
We evaluate the presented approach on the original CARLA benchmark~\cite{dosovitskiy2017carla} (subsequently referred to as \textit{CoRL2017}) and on the recent \textit{NoCrash} benchmark~\cite{codevilla2019exploring}.
At each frame, agents receive a monocular RGB image $I$, velocity $v$, and a high-level command $c$ to compute steering $s$, throttle $t$, and brake $b$, in order to navigate to the specified goals.
Agents are evaluated in an urban driving setting, with intersections and traffic lights.
The \textit{CoRL2017} benchmark consists of four driving conditions, each with $25$ predefined navigation routes.
The four driving conditions are: driving straight, driving with one turn, full navigation with multiple turns, and the same full navigation routes but with traffic.
A trial on a given route is considered successful if the agent reaches the goal within a certain time limit.
The time limit corresponds to the amount of time needed to drive the route at a cruising speed of 10 km/h.
In the \textit{CoRL2017} benchmark, collisions and red light violations do not count as failures. We thus conduct a separate infraction analysis to examine the behavior of the agents in more detail.

The \textit{NoCrash} benchmark~\cite{codevilla2019exploring} consists of three driving conditions, each on a shared set of $25$ predefined routes with comparable difficulty to the full navigation condition in the \textit{CoRL2017} benchmark.
The three conditions differ in the presence of traffic: no traffic, regular traffic, and dense traffic, respectively.
As in \textit{CoRL2017}, a trial is considered successful if the agent reaches the goal within a given time limit. The time limit corresponds to the amount of time needed to drive the route at a cruising speed of 5 km/h.
In addition, in \textit{NoCrash}, a trial is considered a failure if a collision above a preset threshold occurs.

Both benchmarks are evaluated under six weather conditions, four of which were seen during training and the other two only used at test time.
The training weathers are ``Clear noon'', ``Clear noon after rain'', ``Heavy raining noon'', and ``Clear sunset''.
For \textit{CoRL2017}, the test weathers are ``Cloudy noon after rain'' and ``Soft raining sunset'' \cite{codevilla2018end}. For \textit{NoCrash}, the test weathers are ``After rain sunset'' and ``Soft raining sunset''. 
We train a single agent for all conditions.

The CARLA simulator underwent a significant revision in version 0.9.6, including an update of the rendering engine and pedestrian logic.
This makes CARLA 0.9.5 and prior versions not comparable to the current CARLA versions.
We thus compare to all prior work on the older CARLA 0.9.5, but also provide numbers on the newer 0.9.6 version for future reference.
For a fair comparison, we reran the current state-of-the-art CILRS model~\cite{codevilla2019exploring} on CARLA 0.9.5, but were unable to run other methods due to the lack of open implementations or support for newer versions of CARLA.
See the supplement for more detail on the effect of the simulator version on the benchmark.

\mypara{Ablation study.}
Table~\ref{tbl:controlled} compares our full learning-by-cheating (LBC) approach to simpler baselines on the \textit{CoRL2017} benchmark (``navigation'' condition).

\begin{wraptable}{r}{0.5\textwidth}
\centering
\begin{tabular}{lcc c}
\toprule
Supervision & white-box & on-policy \\
\midrule
Direct &  &  & $20$  \\
Two stage & & & $16$  \\
Two stage & & \checkmark & $64$ \\
Two stage & \checkmark & & $96$ \\
Two stage & \checkmark & \checkmark & $100$ \\
\bottomrule
\end{tabular}
\caption{Ablation study on the \emph{CoRL2017} benchmark (CARLA 0.9.5, ``navigation'' condition, test town, test weather). Two key advantages of the presented decomposition~-- white-box supervision and on-policy trajectories~-- each substantially improve performance and together achieve 100\% success rate on the benchmark.}
\label{tbl:controlled}
\end{wraptable}

\begin{table}[b!]
\begin{tabular}{l@{\ }c@{\ \ \ \ }c@{\ \ \ \ }c@{\ \ \ \ }c@{\ \ \ \ }c@{\ \ \ \ }c@{\ \ \ \ }cc}
\toprule
    Task & Weather & MP~\cite{dosovitskiy2017carla} & CIL~\cite{codevilla2018end} & CIRL~\cite{liang2018cirl} & CAL~\cite{Sauer2018CORL} & CILRS~\cite{codevilla2019exploring} & \textbf{LBC} & \textbf{LBC$^\dagger$}\\
    \midrule
    
Straight &\multirow{4}{*}{train}& $92$ & $97$ & $\textbf{100}$ & $93$ & $96$ & $\textbf{100}$ & $\textbf{100}$\\
One turn && $61$ & $59$ & $71$ & $82$ & $84$ & $\textbf{100}$ & $\textbf{100}$\\
Navigation && $24$ & $40$ & $53$ & $70$ & $69$ & $\textbf{100}$ & $\textbf{98}$\\
Nav. dynamic && $24$ & $38$ & $41$ & $64$ & $66$ & $\textbf{99}$ & $\textbf{99}$\\
\midrule
Straight &\multirow{4}{*}{test}& $50$ & $80$ & $98$ & $94$ & $96$ & $\textbf{100}$ & $\textbf{100}$\\
One turn && $50$ & $48$ & $80$ & $72$ & $92$ & $\textbf{100}$ & $\textbf{100}$\\
Navigation && $47$ & $44$ & $68$ & $88$ & $92$ & $\textbf{100}$ & $\textbf{100}$\\
Nav. dynamic && $44$ & $42$ & $62$ & $64$ & $90$ & $\textbf{100}$ & $\textbf{100}$\\
\bottomrule
 \end{tabular}

\vspace{1mm}
\caption{Comparison of the success rate of the presented approach (LBC) to the state of the art on the original CARLA benchmark (\emph{CoRL2017}) in the test town. (The supplement provides results on the training town.) LBC$^\dagger$ denotes our agent trained and evaluated on CARLA 0.9.6. All other agents were evaluated on CARLA 0.8 and 0.9.5. Our approach outperforms all prior work and achieves 100\% success rate on all routes in the full-generalization setting (test town, test weather). 
}
\label{tbl:comparison_legacy}
\vspace{1em}

 \begin{tabular}{l@{\ \ \ \ }c@{\ \ \ \ }c@{\ \ \ \ }c@{\ \ }c@{\ \ }c|c@{\ \ \ \ }c@{\ \ \ \ }c}
\toprule
&& \multicolumn{4}{c|}{CARLA $\le$0.9.5} & \multicolumn{3}{c}{CARLA 0.9.6} \\
Task & Weather & CIL~\cite{codevilla2018end} & CAL~\cite{Sauer2018CORL} & CILRS~\cite{codevilla2019exploring} & LBC & LBC & PV & AT  \\
\midrule
Empty   & \multirow{3}{*}{train}& $48\pm3$ & $36\pm6$ & $51\pm1$ & $\textbf{100}\pm0$ & $100\pm0$ & $100\pm0$ & $100\pm0$ \\
Regular && $27\pm1$ & $26\pm2$ & $44\pm5$ & $\textbf{96}\pm5$ & $94\pm3$ & $95\pm1$ & $99\pm1$ \\
Dense   && $10\pm2$ & $9\pm1$ & $38\pm2$ & $\textbf{89}\pm1$ & $51\pm3$ & $46\pm8$ & $60\pm3$ \\
\midrule
Empty   & \multirow{3}{*}{test}& $24\pm1$ & $25\pm3$ & $90\pm2$ & $\textbf{100}\pm2$ & $70\pm0$ & $100\pm0$ & $100\pm0$ \\
Regular && $13\pm2$ & $14\pm2$ & $87\pm5$ & $\textbf{94}\pm4$ & $62\pm2$ & $93\pm2$ & $99\pm1$ \\
Dense   && $2\pm0$ & $10\pm0$ & $67\pm2$ & $\textbf{85}\pm1$ & $39\pm8$ &$45\pm10$ & $59\pm6$ \\
\bottomrule
 \end{tabular}

\vspace{1mm}
\caption{Comparison of the success rate of the presented approach (LBC) to the previous approaches on the \emph{NoCrash} benchmark in the test town. (The supplement provides results on the training town.) PV denotes the performance of the privileged agent, AT is the performance of the built-in CARLA autopilot. Since the graphics and simulator behavior changed significantly with CARLA 0.9.6, we evaluate and compare our method on CARLA 0.9.5. CILRS was also run on this version of CARLA. Our approach outperforms prior work by significant factors, achieving 100\% success rate in the ``Empty" condition and reaching 85\% success rate or higher in other conditions. 
}
\label{tbl:comparison_nocrash}
\end{table}

``Direct'' one-stage training of the sensorimotor policy by imitation of the autopilot expert does not perform well in our experiments. This is in part due to the lack of trajectory augmentations in our training data. Prior direct imitation approaches~\cite{codevilla2018end,codevilla2019exploring} heavily relied on trajectory noise injection during training.

Vanilla two-stage training suffers from the same issues and does not perform better than a direct supervision.
However, simply supervising all conditional branches during training (``white-box") significantly increases the performance of the model.
White-box supervision appears to function as powerful data augmentation that is extremely effective even in the off-policy setting.
Consider an intersection with left and right turns. In traditional imitation learning, the student only receives gradients for the branch taken by the supervising agent, and only on the supervising agent's near-perfect trajectory. With white-box supervision, the sensorimotor student receives supervision on what it would need to do even if it suddenly had to turn right in the middle of a left turn.
Multi-branch training also helps the sensorimotor model decorrelate its outputs across branches, as it gets to see multiple different signals for each training example.

``On-policy" refers to the sensorimotor agent rolling out its own policy during training.
Training with both white-box multi-branch supervision and student rollouts (bottom row in Table~\ref{tbl:controlled}) yields the best results and achieves 100\% success rate in all conditions.
We use this setting in all experiments that follow.

\mypara{Comparison to the state of the art.}
\lblsec{quantitative_results}
Tables~\ref{tbl:comparison_legacy} and~\ref{tbl:comparison_nocrash} compare the performance of our final sensorimotor agent to the state of the art on the \textit{CoRL2017}~\cite{dosovitskiy2017carla} and \textit{NoCrash}~\cite{codevilla2019exploring} benchmarks, respectively. We substantially outperform the prior state of the art on both benchmarks. On \textit{CoRL2017}, we achieve 100\% success rate on all routes in the full-generalization setting (new town, new weather). On \textit{NoCrash}, we outperform the recent CILRS model~\cite{codevilla2019exploring} by significant factors, achieving 100\% success rate without traffic and reaching 85\% success rate or higher in all conditions.

\begin{wrapfigure}{r}{0.6\textwidth}
\vspace{-2mm}
\centering
\begin{subfigure}[b]{0.5\linewidth}
\begin{tikzpicture}
\begin{axis}[
    barplot,
    ylabel={\scriptsize infractions per 10km},
    ylabel near ticks,
    clip=false,
    minor grid style={dotted, aluminium3},
    symbolic x coords={T1, T1S, T2, T2S},
    xticklabels={,T1, T1*, T2, T2*},
    enlarge x limits=0.2,
    ymin=0,ymax=80,
	width=\linewidth,
	bar width=0.6em,
nodes near coords,
every node near coord/.append style={black,font=\scriptsize,/pgf/number format/.cd,precision=0,},
]   
\addplot[scarletred3,fill,error bars/.cd, y dir=both, y explicit, error bar style={black,ultra thick}] coordinates {
(T1,37.893) += (0,1.103) -= (0, 1.103)
(T1S,40.551) += (0,1.170) -= (0, 1.170)
(T2,78.893) += (0,4.074) -= (0, 4.074)
(T2S,61.738) += (0,3.495) -= (0, 3.495)};
\addplot[skyblue3,fill,error bars/.cd, y dir=both, y explicit, error bar style={black,ultra thick}] coordinates {
(T1,2.584) += (0,0.755) -= (0,0.755)
(T1S,2.516) += (0,0.288) -= (0,0.288)
(T2,3.346) += (0,0.205) -= (0,0.205)
(T2S,4.554) += (0,0.094) -= (0,0.094)};

\end{axis}
\end{tikzpicture}
\caption{Traffic light violations}
\end{subfigure}%
\begin{subfigure}[b]{0.5\linewidth}
\begin{tikzpicture}
\begin{axis}[
    barplot,
    ylabel={\scriptsize infractions per 10km},
    ylabel near ticks,
    clip=false,
    minor grid style={dotted, aluminium3},
    symbolic x coords={T1, T1S, T2, T2S},
    xticklabels={,T1, T1*, T2, T2*},
    enlarge x limits=0.2,
    ymin=0,ymax=200,
	width=\linewidth,
	bar width=0.6em,
nodes near coords,
every node near coord/.append style={black,font=\scriptsize,/pgf/number format/.cd,precision=0,},
		legend columns=2,
    	legend style={at={(0.95,0.85)},font=\scriptsize},
]   
\addplot[scarletred3,fill,error bars/.cd, y dir=both, y explicit, error bar style={black,ultra thick}] coordinates {
(T1,30.8967) += (0,0.596) -= (0, 0.596)
(T1S,28.4567) += (0,1.562) -= (0, 1.562)
(T2,209.1433) += (0,22.109) -= (0, 22.109)
(T2S,56.4733) += (0,1.265) -= (0, 1.265)};
\addplot[skyblue3,fill,error bars/.cd, y dir=both, y explicit, error bar style={black,ultra thick}] coordinates {
(T1,3.0000) += (0,1.460) -= (0,1.460)
(T1S,1.5900) += (0,1.187) -= (0,1.187)
(T2,12.1667) += (0,1.498) -= (0,1.498)
(T2S,17.7100) += (0,4.075) -= (0,4.075)};
\legend{CILRS, LBC}
\end{axis}
\end{tikzpicture}
\caption{Collisions}
\end{subfigure}%
\caption{Infraction analysis. Number of traffic light violations and crashes per $10$ km in Town 1 (T1), Town 1 with new weather (T1*), Town 2 (T2), and Town 2 with new weather (T2*).}
\lblfig{infraction_analysis}
\end{wrapfigure}
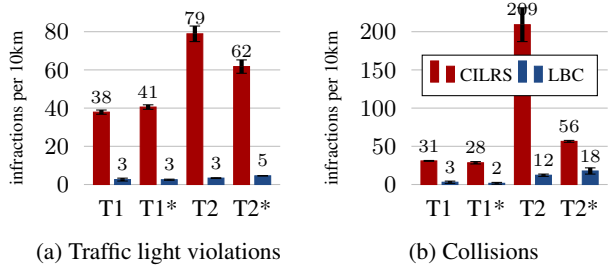
\mypara{Infraction analysis.}
To examine the driving behavior of the agents in further detail, we conduct an infraction analysis on all routes from the \textit{NoCrash} benchmark in CARLA 0.9.5. We compare the presented approach (LBC) with the previous state of the art (CILRS~\cite{codevilla2019exploring}). We measure the average number of traffic light violations (i.e., running a red light) and collisions per 10 km. The results are summarized in \reffig{infraction_analysis}. Our approach cuts the frequency of infractions by at least an order of magnitude in most conditions.

\section{Conclusion}
\label{sec:conclusion}

We showed that imitation learning for vision-based urban driving can be made much more effective by decomposing the learning process into two stages: first training a privileged (``cheating") agent and then using this privileged agent as a teacher to train a purely vision-based system. This decomposition partially decouples learning to act from learning to see and has a number of advantages. We have validated these advantages experimentally and have used the presented approach to train a vision-based urban driving system that substantially outperforms the state of the art on standard benchmarks.

Our training procedure leverages simulation, and indeed highlights certain benefits of simulation. (``Cheating" by accessing the ground-truth state of the environment is difficult in the physical world.) However, the procedure yields genuine vision-based driving systems that are not tied to simulation in any way. They can be transferred to the physical world using any procedure for sim-to-real transfer~\cite{muller2018driving,bewley2018learning}. We leave such demonstration to future work and hope that the presented ideas will serve as a powerful shortcut en route to safe and robust autonomous driving systems.

Another exciting opportunity for future work is to combine the presented ideas with reinforcement learning and train systems that exceed the capabilities of the expert that provides the initial demonstrations~\cite{Sun2017}.

Our implementation and benchmark results are available at \url{https://github.com/dianchen96/LearningByCheating}.


\acknowledgments{We acknowledge the Texas Advanced Computing Center (TACC) for providing computing resources. This work has been supported in part by the National Science Foundation under grants IIS-1845485 and CNS-1414082. We thank Xingyi Zhou for constructive feedback on the network architecture, and Felipe Codevilla for image augmentation code.}

\bibliography{corl_2019}  

\begin{appendices}
\section{Additional details}

\mypara{Map-view perspective transformation.}
Given a predicted waypoint $\mathbf{\tilde w}=(\tilde{w}_x, \tilde{w}_y)$ in camera coordinates, we compute its projection onto the ground plane $(\hat w_x,\hat w_y,0)$ using the camera's horizontal field of view (fov) $f = \frac{w}{2\tan(fov/2)}$, height $p_y$, and canvas center $c_x=\frac{w}{2}, c_y=\frac{h}{2}$.
We assume the camera always faces forward, as the map is anchored at the agent's position and local coordinate frame.
We always project points onto a constant ground plane $z=0$ to avoid depth estimation:
$
    \hat w_y = \frac{f}{c_y-\tilde{w}_y} p_y,
    \hat w_x = \frac{\tilde{w}_x-c_x}{c_y-\tilde{w}_y}p_y
$.
To make the projection a one-to-one mapping, we additionally move the projected points back by $4$ meters, to prevent points closer to the ego-vehicle getting clipped at the bottom of the image. This transformation is differentiable and the sensorimotor agent can be trained end-to-end. The camera is placed at $p=(2,0,1.4)$ in the vehicle's reference frame, at the hood position. The camera faces forward and has a resolution of $384 \times 160$ with horizontal fov $90^\circ{}$. 

\mypara{Data collection.}
To train the privileged agent, we use 157K training frames and 39K validation frames collected at 10 fps by a hand-crafted autopilot. We use 174K training frames in our 0.9.6 implementation. For both our offline dataset collection and privileged rollouts, we collect the frames using four training weather conditions uniformly sampled in the training town. We add $100$ other vehicles to share the traffic with the ego-vehicle. We add $250$ pedestrians in our 0.9.6 implementation.

\mypara{Hyperparameters.}
We use the Adam optimizer with initial learning rate $10^{-4}$ and no weight decay to train all our models. We use batch size $32$ to train all of our models in 0.9.5, batch size $128$ to train the privileged model in 0.9.6, and batch size $96$ to train the sensorimotor model in 0.9.6. We used the batch augmentation trick~\cite{hoffer2019augment} with $m=4$ for our image model training in 0.9.6. For the spatial argmax layers in both privileged and sensorimotor agent, we fix temperature $\beta=1$ instead of a learnable parameter.  We use PyTorch 1.0 to train and evaluate our models.

\mypara{Image model warm-up.}
Since a randomly initialized network returns the canvas center at the end of the spatial argmax layer in the image coordinate, it corresponds to infinite distance when projected. This causes gradients to explode in the backward pass. To address this issue, we warm up our image model by first supervising it with loss in the projected image coordinate space for 1K iterations before the two-stage training.

\section{Additional experiments}

If an agent can perform near perfectly using a map representation, why not simply try to predict the map representation from raw pixels, then act on that?
This approach resembles that of~\citet{muller2018driving}, where perception and control are explicitly decoupled and trained separately.

We train two networks. The first network is used for perception and directly predicts the privileged representation from an RGB image.
We resize the RGB image to $192 \times 192$ and feed this into a ResNet34 backbone~\cite{he2016deep}, followed by five layers of bilinear-upsampling + convolution + ReLU to produce a map of the original resolution of $192 \times 192$.
The perception network is trained using an L1 loss between the network's output, and the ground truth privileged representation.
The second network is used for action and predicts waypoints from the output of the first network.
We use the same architecture and training procedure as the privileged agent in our main experiments, and we freeze the weights of the perception network during training.

We use a dataset of 150K frames collected from an expert in training conditions, with no trajectory noise.
The offline map predictions on the training and validation sets are quite good, but we notice that during evaluation, even slightly out-of-distribution observations produce erroneous map predictions, causing the waypoint network to fail.
To address this, we collect another dataset of the same size and employ trajectory noise~\cite{codevilla2018end} in 20\% of the frames, to broaden the states seen by the perception network.
\reftbl{map_pred} shows the results.
Both map prediction agents performs significantly worse than our two-stage agent.

\begin{table*}[h]
\centering
\begin{tabular}{l cc cc}
\toprule
& \multicolumn{2}{c}{Train Town} & \multicolumn{2}{c}{Test Town} \\
Method &  Train Weather & Test Weather & Train Weather & Test Weather \\
\midrule
No augmentation & $39$ & $34$ & $30$ & $32$  \\
Trajectory noise & $58$ & $62$ & $65$ & $62$ \\
\midrule
LBC & $100$ & $100$ & $100$ & $100$ \\
\bottomrule
\end{tabular}
\caption{Map prediction baseline with and without trajectory noise compared to our two-stage LBC, evaluated on the Navigation task of the \textit{CoRL2017} benchmark on CARLA 0.9.5.
}
\lbltbl{map_pred}
\end{table*}

\section{Benchmark results}
For completeness, \reftbl{comparison_legacy_train} and \reftbl{comparison_nocrash} show the training town performance in the CARLA CoRL 2017 and NoCrash benchmarks, respectively.
We again compare to MP~\cite{dosovitskiy2017carla}, CIL~\cite{dosovitskiy2017carla, codevilla2018end}, CAL~\cite{Sauer2018CORL}, CIRL~\cite{liang2018cirl}, and CILRS~\cite{codevilla2019exploring}.

\reftbl{compare_carla_cilrs} compares different CARLA versions, 0.8 and 0.9.5.
We compare the performance of CILRS on the Navigation Dynamic task with and without a working pedestrian autopilot (versions 0.8 and 0.9.5, respectively).
The CILRS performance in 0.9.5 matches the older CARLA version in test weathers and is slightly lower in the training weathers. This indicates that CARLA 0.9.5 does not make the task easier.
We report the higher numbers from the CILRS paper~\cite{codevilla2019exploring}.

\begin{table*}[ht!]

\centering
{
\resizebox{1.0\linewidth}{!}
{

\begin{tabular}{l@{\ }c@{\ \ \ \ }c@{\ \ \ \ }c@{\ \ \ \ }c@{\ \ \ \ }c@{\ \ \ \ }c@{\ \ \ \ }cc}
\toprule
    Task & Weather & MP\cite{dosovitskiy2017carla} & CIL\cite{codevilla2018end} & CIRL\cite{liang2018cirl} & CAL\cite{Sauer2018CORL} & CILRS\cite{codevilla2019exploring} & \textbf{LBC} & \textbf{LBC$^\dagger$}\\
    \midrule
    
Straight &\multirow{4}{*}{train}& $98$ & $98$ & $98$ & $\textbf{100}$ & $96$ & $\textbf{100}$ & $\textbf{100}$ \\
One Turn && $82$ & $89$ & $97$ & $97$ & $92$ & $\textbf{100}$ & $\textbf{100}$\\
Navigation && $80$ & $86$ & $93$ & $92$ & $95$ & $\textbf{100}$ & $\textbf{100}$\\
Nav. Dynamic && $77$ & $83$ & $82$ & $83$ & $92$ & $\textbf{100}$ & $\textbf{100}$\\
\midrule
Straight &\multirow{4}{*}{test}& $\textbf{100}$ & $98$ & $\textbf{100}$ & $\textbf{100}$ & $96$ & $\textbf{100}$ & $\textbf{100}$\\
One Turn && $95$ & $90$ & $94$ & $\textbf{96}$ & $\textbf{96}$ & $\textbf{100}$ & $\textbf{96}$\\
Navigation && $94$ & $84$ & $86$ & $90$ & $96$ & $\textbf{100}$ & $\textbf{100}$\\
Nav. Dynamic && $89$ & $82$ & $80$ & $82$ & $\textbf{96}$ & $\textbf{96}$ & $\textbf{96}$ \\
\bottomrule
 \end{tabular}
}
}
\caption{Quantitative results on the training town in the CoRL2017 CARLA benchmark. LBC$^\dagger$ denotes our agent trained and evaluated on our customized CARLA based on 0.9.6, the most up-to-date CARLA version. Note that CARLA 0.9.6 has different graphics compared to 0.8 and 0.9.5.}
\lbltbl{comparison_legacy_train}
\end{table*}

\lblsec{comparison_cilrs}
\begin{table*}
\centering
 \begin{tabular}{l@{\ \ \ \ }c@{\ \ \ \ }c@{\ \ \ \ }c@{\ \ }c@{\ \ }c|c@{\ \ \ \ }c@{\ \ \ \ }c}
\toprule
&& \multicolumn{4}{c|}{CARLA $\le$0.9.5} & \multicolumn{3}{c}{CARLA 0.9.6} \\
Task & Weather & CIL\cite{codevilla2018end} & CAL\cite{Sauer2018CORL} & CILRS\cite{codevilla2019exploring} & LBC & LBC & PV & AT  \\
\midrule
Empty   & \multirow{3}{*}{train}& $79\pm1$ & $81\pm1$ & $87\pm1$ & $\textbf{100}\pm0$ & $\textbf{97}\pm1$ & $100\pm1$ & $100\pm0$ \\
Regular && $60\pm1$ & $73\pm2$ & $83\pm0$ & $\textbf{99}\pm1$ & $\textbf{93}\pm1$ & $96\pm3$ & $99\pm1$ \\
Dense   && $21\pm2$ & $42\pm1$ & $42\pm2$ & $\textbf{95}\pm2$ & $\textbf{71}\pm5$ & $80\pm5$ & $86\pm3$ \\
\midrule
Empty   & \multirow{3}{*}{test}& $83\pm2$ & $85\pm2$ & $87\pm1$ & $\textbf{100}\pm0$ & $87\pm4$ & $100\pm0$ & $100\pm0$ \\
Regular && $55\pm5$ & $68\pm5$ & $88\pm2$ & $\textbf{99}\pm1$ & $\textbf{87}\pm3$ & $97\pm3$ & $99\pm1$ \\
Dense   && $13\pm4$ & $33\pm2$ & $70\pm3$ & $\textbf{97}\pm2$ & $\textbf{63}\pm1$ & $81\pm6$ & $83\pm6$ \\
\bottomrule
 \end{tabular}
\caption{Quantitative results on the training town in the NoCrash benchmark. The methods were run on CARLA $<=$0.9.5. LBC$^\dagger$ denotes our agent trained and evaluated on our customized CARLA based on 0.9.6, the most up-to-date CARLA version. Note that CARLA 0.9.6 has different graphics compared to 0.8 and 0.9.5.}
\lbltbl{comparison_nocrash}
\end{table*}

\begin{table*}[ht!]
\begin{tabular}{ll cc cc}
\toprule
& & \multicolumn{2}{c}{Train Town} & \multicolumn{2}{c}{Test Town} \\
& CARLA version & Train Weather & Test Weather & Train Weather & Test Weather \\
\midrule
CILRS~\cite{codevilla2019exploring} & 0.8.4 & $92$ & $96$ & $66$ & $90$ \\
CILRS & 0.9.5 (no ped) & $84$ & $96$ & $53$ & $92$ \\
\bottomrule
\end{tabular}
\caption{Comparison of CILRS~\cite{codevilla2019exploring} on different versions of CARLA on the Navigation Dynamic task of the CoRL2017 benchmark.} 
\lbltbl{compare_carla_cilrs}
\end{table*}

\end{appendices}


\end{document}